# Modeling Large-Scale Walking and Cycling Networks: A Machine Learning Approach Using Mobile Phone and Crowdsourced Data

Meead Saberi[1,*] and Tanapon Lilasathapornkit[1]

[1]*School of Civil and Environmental Engineering, Research Centre for Integrated Transport Innovation (rCITI), University of New South Wales, Sydney, Australia.*

\* *Corresponding author (meead.saberi@unsw.edu.au)*

## ABSTRACT

Walking and cycling are known to bring substantial health, environmental, and economic advantages. However, the development of evidence-based active transportation planning and policies has been impeded by significant data limitations, such as biases in crowdsourced data and representativeness issues of mobile phone data. In this study, we develop and apply a machine learning based modeling approach for estimating daily walking and cycling volumes across a large-scale regional network in New South Wales, Australia that includes 188,999 walking links and 114,885 cycling links. The modeling methodology leverages crowdsourced and mobile phone data as well as a range of other datasets on population, land use, topography, climate, etc. The study discusses the unique challenges and limitations related to all three aspects of model training, testing, and inference given the large geographical extent of the modeled networks and relative scarcity of observed walking and cycling count data. The study also proposes a new technique to identify model estimate outliers and to mitigate their impact. Overall, the study provides a valuable resource for transportation modelers, policymakers and urban planners seeking to enhance active transportation infrastructure planning and policies with advanced emerging data-driven modeling methodologies.

## 1. INTRODUCTION

There is a growing global interest in active transport strategic planning and policies due to their health, environmental, and economic benefits. However, the lack of quality active transportation data has been hindering informed policy and infrastructure investment decision-making. Crowdsourced data such as Strava can capture cycling patterns, but they have been shown to be biased towards specific types of travelers (Garber et al., 2019; Roy et al., 2019; Venter et al., 2023). Mobile phone data can also capture active travel patterns, but the sample representativeness and the quality of the data, especially for short trips such as walking and cycling, have been subjects of debate (Calabrese et al., 2013; Lu et al, 2017; Conrow et al., 2018; Caceres et al., 2020; Sinclair et al., 2023). A growing number of studies have proposed methods to correct the sampling bias and inaccuracies in crowdsourced and mobile phone data in the context of active transport planning (Jestico et al., 2016; Chen et al., 2020; Nelson et al., 2021a). Integrating the emerging crowdsourced and mobile phone data with other data sources have been proven to estimate more representative active mobility patterns. This study aims to develop a data-driven machine learning-based active transportation modeling approach including all three steps of model training, testing and inference to estimate link-level walking and cycling volumes across a large-scale area in New South Wales (NSW), Australia using a wide range of data sources including observed walking and cycling counts, crowdsourced data, mobile phone data, population and land use, topography, climate, air quality, household travel survey, etc.

Crowdsourced data often lack the quality assurance of traditional geographic data collection measures. Jestico et al. (2016) raised concerns around potential biases from user submitted content that are difficult to quantify without comparing against reference data sources. They analyzed Strava cycling data in the Vancouver, BC, Canada area and found that the average Strava sampling to population ratio is 1:51 with significant variation across different sites ranging from 10% error to 100% error. In a follow up study, Roy et al. (2019) argued that crowdsourced cycling data are biased as they oversample recreational riders. However, they demonstrated that different geographical variables can be quantified to correct the bias in crowdsourced data. They trained a model using observed cycling counts from 44 locations across the Maricopa Association of Governments (MAG), Arizona, USA. They demonstrated that the proposed modeling approach is broadly applicable for correcting bias in crowdsourced active transportation data when official observed counts and geographical data are available. In a more recent study, Nelson et al. (2021a) also argued that to overcome the bias in crowdsourced data, statistical models can be developed to estimate total cycling volumes by integrating Strava data with official observed counts and a range of geographic data. They developed and tested the modeling approach in five different cities across North America including Boulder (CO), Ottawa (ON), Phoenix (AZ), San Francisco (CA), and Victoria (BC).

Despite the growing literature on the use of emerging data sources and modeling methodologies to overcome the bias in the data, most previous studies focused on cycling only and demonstrated the applicability of the modeling approach in relatively small-scale networks and were limited to use of observed count data from small number of locations. The presented machine learning-based modeling approach in this study, building upon the previous literature, extends the methodology to walking and successfully demonstrates the applicability, quality, and validity of the proposed machine-learning approach through training and testing of large-scale regional walking and cycling volume estimation models for the New South Wales (NSW) Six Cities Region in Australia. To the best of authors' knowledge, the developed models with 188,999 walking links and 114,885 cycling links are the largest and finest in resolution models of their kind ever developed and documented in the literature to date. The study also provides new insights into the unique challenges and limitations of large-scale model inference and proposes new techniques to identify model estimate outliers and to mitigate their impact.

The contributions of the study lie in the introduction of a comprehensive machine learning-based modeling approach that integrates crowdsourced and mobile phone data with diverse datasets, overcoming data limitations in active transportation planning. The development of models with an unprecedented scale addresses a crucial gap in previous research, allowing for estimation of walking and cycling volumes across large-scale networks. The study applies various techniques to address biases in crowdsourced and mobile phone data, enhancing the reliability and validity of the machine learning models in estimating active mobility patterns. Unlike previous studies focusing mainly on cycling, this study extends the methodology to walking, showcasing its applicability, quality, and validity, providing valuable insights for active transportation planning and management.

The remainder of the manuscript is organized as follows. Section 2 provides a comprehensive description of the data used in this study. Section 3 describes the modeling approach and results including feature importance analysis and selection, model training, and model testing. Section 4 describes application of the model to the large-scale NSW Six Cities Region and discusses model inference challenges. Section 5 provides a discussion on the study limitations and future research directions.

## 2. DATA

### 2.1. Study area

The NSW Six Cites Region encompasses six cities in the state of NSW, Australia including Lower Hunter and Greater Newcastle, Central Coast, Illawarra-Shoalhaven, Western Parkland, Central River, and Eastern Harbour. The region encompasses 43 local government areas with a total area of over 2.25 million hectares and population of 6.27 million residents (2021).

## 2.2. Walking and cycling networks

The walking network of the study area is constructed using the Geoscape PSMA network (Geoscape Australia, 2023) including a total of 188,999 links. The cycling network is constructed using OpenStreetMap and Transport for NSW Bicycle Infrastructure Network data (Transport for NSW, 2023) including a total of 114,885 links. See Figure 1. The large geographical extent of the two networks introduces unique modeling challenges related to all three aspects of model training, testing, and inference that are discussed later in the manuscript in details and shape the main contributions of this study.

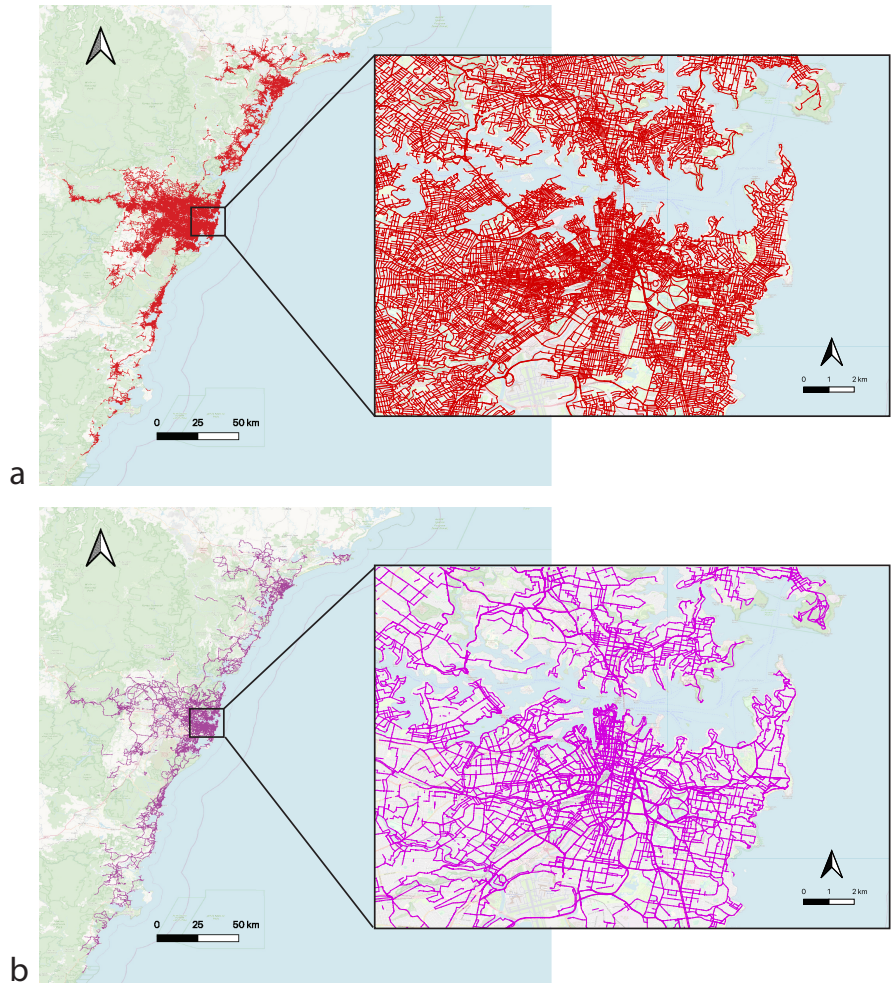

**Figure 1.** Geographical extent of the constructed (a) walking with 188,999 links and (b) cycling network with 114,885 links in the NSW Six Cities Region, Australia.

## 2.3. Observed walking and cycling counts

Official observed walking and cycling count data are obtained from a range of sources including Transport for NSW open data hub[1], City of Sydney open data hub[2], and City of Parramatta Pedestrian and Cyclist Dashboard[3] among others. A total of 27,631 walking count records and 18,535 cycling count records between 2013 and 2023 across 766 sites are collated. See Figure 2. The spatial diversity of the observed count locations is of outmost importance in the presented modeling approach and can have significant impacts on the quality of the modeling outcomes, which will be discussed in detail in section 5 and 6 of the manuscript.

---

[1] https://opendata.transport.nsw.gov.au/dataset/collective-bicycling -datasets
[2] https://data.cityofsydney.nsw.gov.au/
[3] https://www.cityofparramatta.nsw.gov.au/recreation/walking-and-bicycling /pedestrian-and-cyclist-dashboard

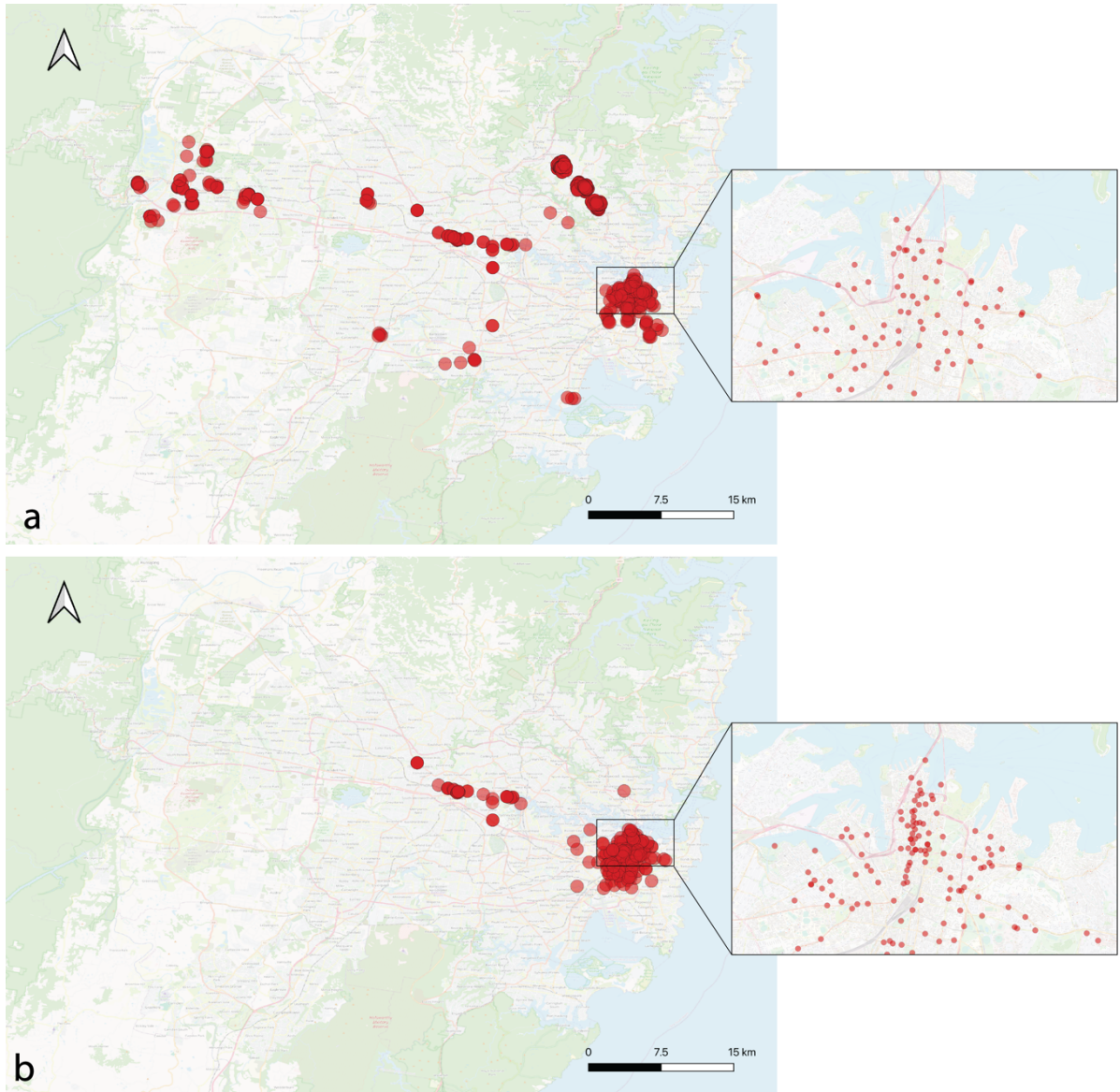

**Figure 2.** Spatial distribution of the (a) 634 walking and (b) 160 cycling observed count locations across the study area.

### 2.4. Mobile phone and crowdsourced data

We use link-level estimates of walking counts from a private telecommunications company that takes billions of data points generated daily by mobile phone devices across Australia, cleans, aggregates, and anonymizes it using their proprietary analytics framework to generate mobility insights including aggregated and link-level estimates of walking trips. We also use link-level cycling counts from Strava Metro[4]. Strava Metro derives data from Strava fitness app's GPS records and offers both aggregated and link-level cycling counts. Both data types have been widely used and explored in the literature with well-documented issues in relation to sampling bias, localization error, etc. (Wesolowski et al., 2013; Calabrese et al., 2013; Cacares et al., 2020; Nelson et al., 2021b).

---

[4] https://metro.strava.com/

## 2.5. Population and land use data

We use population and land use data from Australian Bureau of Statistics (ABS) Census 2021 at the Statistical Areas Level 1 (SA1s) level across the study area. SA1s are geographic areas in Australia created from whole Mesh Blocks, totaling 61,845 zones without gaps or overlaps. They are designed to maximize geographic detail for Census data, with each SA1 having a population of 200 to 800 people, representing urban or rural characteristics. Here, we specifically work with the following population and land use data that are shown to have significant association with the bias in the crowdsourced and mobile phone data in the context of walking and cycling (Roy et al., 2019; Nelson et al., 2021a):

1. Population density (person/sq.km)
2. Person weekly median income (AUD)
3. Land use mix entropy

Land use mix entropy (LUM) is a measure of the heterogeneity/homogeneity of land uses which uses the areas of different land use categories within each SA1 polygon to calculate an index between 0 and 1. An index of 1 represents a completely mixed land use while an index of 0 represents a single use land within the SA1. LUM is calculated as follows (Christian et al., 2011).

$$LUM = -1\left(\sum_{i=1}^{n} p_i * ln(p_i)/ln(n)\right) \quad (1)$$

where $p_i$ is the proportion of the SA1 covered by the land use class $i$ against the summed area for all land use classes and $n$ in the number of land use classes. Here, we use all available land use classes in the ABS Census data including residential, industrial, education, parkland, and hospital/medical.

## 2.6. Household Travel Survey (HTS) data

The NSW HTS[5] is a continuous data collection initiative started in 1997/98 that focuses on personal travel behavior within the Sydney Greater Metropolitan Area. It involves randomly selecting residents from occupied private dwellings in the survey area, with approximately 2,000-3,000 households participating annually. The survey collects data on all trips including active transportation made over a 24-hour period by members of participating households, producing annual estimates weighted to the ABS' estimated resident population for an average weekday.

## 2.7. Air quality, climate and topography data

Air quality data including PM2.5, PM10 and visibility (NEPH) information are obtained from NSW Government Department of Planning, Industry and Environment portal[6]. Climate data including precipitation and temperature are obtained from the Australian Bureau of Meteorology[7]. Each link in the study networks is then linked with the closest weather station and air quality station to obtain precipitation (mm), temperature (C), PM2.5 (μg/m3) and other air quality metrics. Topography information is obtained from Geoscience Australia, Digital Elevation Map portal[8] with which we estimate an average and maximum slope for each link in the study networks.

See Table 1 for a summary of all variables, their associated spatial and temporal aggregation, and data sources used in this study.

---

[5] https://www.transport.nsw.gov.au/data-and-research/data-and-insights/surveys/household-travel-survey-hts
[6] https://www.dpie.nsw.gov.au/air-quality/air-quality-concentration-data-updated-hourly
[7] http://www.bom.gov.au/climate/data/
[8] http://www.ga.gov.au/scientific-topics/national-location-information/digital-elevation-data

**Table 1.** Summary of all variables and data sources considered in this study.

| Category | Variables | Spatial (and temporal) aggregation | Data source |
|---|---|---|---|
| Bicycling | Observed cycling count data representing Average Daily Bicycling Volume (ADBV) *(response variable)* | Link level (daily/sub-daily) | Obtained from local governments across the study area |
| | Strava bicycle count | Link level (daily) | Strava Metro[9] |
| Walking | Official walking count data representing Average Daily Pedestrian Volume (ADPV) *(response variable)* | Link level (daily/sub-daily) | Obtained from local governments across the study area |
| | Mobile phone-based walking count | Link level (daily) | Private telecommunications company |
| Bicycle infrastructure | Dummy variable for dedicated bicycle paths (1 if there is a dedicated bicycle path with "bicycle only" facility tag, 0 otherwise) | Link level | TfNSW Bicycle Network[10] |
| Walking infrastructure | Dummy variable for residential areas (1 if the link has "residential" on "highway" tag, 0 otherwise) Dummy variable for pedestrian only paths (1 if the link has "footway" on "highway" tag, 0 otherwise) Dummy variable representing roads connecting smaller settlements in an urban area (1 if the link has "tertiary" on "highway" tag, 0 otherwise) | Link level | OpenStreetMap (OSM) |
| Population | Population density (person/sq.km) | SA1 | ABS Census 2016 and 2021 |
| | Person weekly median income (AUD) | SA1 | ABS Census 2016 and 2021 |
| Land use | Land use mix entropy | SA1 | ABS Census 2016 and 2021 |
| | Percentage of parkland land use (%) | SA1 | |
| | Number of Place of Interests (POIs) per area (/sq.km) | LGA | OpenStreetMap (OSM) |
| Air quality | PM2.5 (µg/m³) and PM10 (µg/m³) NEPH[11] NO (ppm), $NO_2$ (pphm), and $O_3$ (pphm) Dummy variable for poor air quality (1 if PM2.5 exceeds 10, 0 otherwise) Dummy variable for low visibility (1 if NEPH is less than 2.3, 0 otherwise) | Catchments of air quality stations (daily/sub-daily) | NSW Department of Planning and Environment |

---

[9] https://metro.strava.com/

[10] https://opendata.transport.nsw.gov.au/dataset/infrastructure-cycleway-data

[11] NEPH represents measurements reported by a nephelometer, as a measure of light scattering or reduction due to atmospheric particulate matter (PM). Scattering by PM impairs visibility, therefore this parameter is also referred to as visibility, as it indicates how visual range is affected by airborne particulate matter.

**Table 1.** (continued)

| Category | Variables | Spatial (and temporal) aggregation | Data source |
|---|---|---|---|
| Climate | Precipitation (mm)<br>Maximum Temperature (C)<br>Minimum Temperature (C)<br>Dummy variable for hot days (1 if temperature exceeds 30C, 0 otherwise)<br>Dummy variable for cold days (1 if temperature is less than 10C, 0 otherwise)<br>Dummy variable for rainy days (1 if the cumulative daily precipitation exceeds 50mm, 0 otherwise) | Catchments of weather stations (daily/sub-daily) | Bureau of Meteorology |
| Topography | Maximum slope (%)<br>Average slope (%)<br>A dummy variable identifying links that are steeper than a threshold (1 if the maximum slope is less than 7.5%, 0 otherwise) | Link level | Digital Elevation Data (DEM) from GeoScience Australia |
| Household Travel Survey | Percentage of "walking only" trips[12]<br>Percentage of "linked walking" trips[13] | LGA | NSW HTS data |
| COVID-19 restrictions | Dummy variable representing the period before COVID-19 (1 if the date is before 1 February 2020, 0 otherwise)<br>Dummy variable representing the NSW lockdown periods (1 if the date is between 16 March 2020 and 15 May 2020 (first wave) or between 23 June 2021 and 15 September 2021 (second wave), 0 otherwise) | Daily | NSW Health |
| Day of week | Dummy variable (1 represents weekday, 0 otherwise) | Daily | N/A |

---

[12] "Walking only" trips are trips where the whole trip is made by walking and no change of mode is involved.

[13] "Linked walking" trips are trips where the purpose is access to, or egress from, another mode e.g., walk to the bus stop to catch the bus or walk from the bus stop after getting off the bus at the other end.

## 2.8. Descriptive statistics of the input data

Tables 2 and 3 provide further details on the quantity, temporal coverage and descriptive statistics of the input data used in development of the walking and cycling models, respectively. In Table 2, focusing on the walking model, various datasets are outlined, including official observed walking counts, mobile phone-based walking counts as well as some of the climate and air quality metrics. The number of records, temporal coverage (from and to dates), and data sources are specified for each category, providing essential details for better understanding the input data. Table 3, dedicated to the cycling model, follows a similar structure, featuring observed cycling counts, Strava cycling counts, as well as data on precipitation, temperature, and a few of the air quality metrics, with corresponding records, temporal coverage, and additional notes on number of associated weather and air quality stations that the data were collated from.

**Table 2.** Number of data records and temporal coverage of the data used in the walking model.

| Data | Number of Records | From | To | Note |
|---|---|---|---|---|
| Observed walking counts *(response variable)* | 27,631 | 01/10/2013 | 28/03/2023 | Collated from local and state government sources |
| Mobile phone-based walking counts | 95,661,501 | 01/01/2019 | 07/07/2022 | Available from two periods between 1/2019 – 5/2019 and between 12/2021 – 7/2022 |
| Precipitation | 462,326 | 01/01/1862 | 10/05/2023 | Across 35 weather stations |
| Temperature | 328,989 | 01/01/1939 | 10/05/2023 | Across 34 weather stations |
| PM2.5 | 9,763 | 07/01/2019 | 05/07/2022 | Across 12 air quality stations |
| PM10 | 9,900 | 07/01/2019 | 05/07/2022 | Across 12 air quality stations |
| NEPH | 10,041 | 07/01/2019 | 05/07/2022 | Across 12 air quality stations |

**Table 3.** Number of data records and temporal coverage of the data used in the cycling model.

| Data | Number of Records | From | To | Notes |
|---|---|---|---|---|
| Observed cycling counts *(response variable)* | 18,535 | 07/01/2019 | 08/08/2021 | Collated from local and state government sources |
| Strava cycling counts | 50,027,617 | 01/01/2019 | 31/07/2021 | Total ridership including both commute and leisure trips |
| Precipitation | 13,636 | 01/01/2019 | 31/08/2021 | Across 13 weather stations |
| Temperature | 8,766 | 01/01/2019 | 31/08/2021 | Across 8 weather stations |
| PM2.5 | 29,634 | 01/01/2018 | 08/09/2021 | Across 21 air quality stations |
| PM10 | 32,328 | 01/01/2018 | 08/09/2021 | Across 23 air quality stations |
| NEPH | 32,328 | 01/01/2018 | 08/09/2021 | Across 23 air quality stations |

Tables 4 and 5 provide further descriptive statistics (minimum, mean, median, maximum, and standard deviation) of the input data for the walking and cycling models, respectively. The largest observed walking and cycling count across the study area was 194,400 pedestrians per day and 1,358 cyclists per day, respectively. The mean observed walking and cycling count was 4,823 pedestrians per day and 188 cyclists per day. The statistics suggest that the majority of the observed count data are located in relatively high pedestrian and cycling activity areas that may impact the quality of the model training and testing that will be further discussed later in the manuscript.

**Table 4.** Descriptive statistics of the input data for the walking model.

| Data | Min | Mean | Median | Max | Std. deviation |
| --- | --- | --- | --- | --- | --- |
| Observed walking counts | 0.00 | 4,822.73 | 616.00 | 194,400 | 12,884.15 |
| Mobile phone-based walking counts | 40.00 | 1,801 | 351.0 | 166,031 | 6,167.02 |
| Precipitation (mm) | 0.00 | 2.88 | 0.00 | 309.40 | 9.56 |
| Minimum Temperature (C) | -7.20 | 12.40 | 12.70 | 29.70 | 5.30 |
| Maximum Temperature (C) | 0.00 | 22.71 | 22.30 | 48.90 | 5.42 |
| PM2.5 ($\mu g/m^3$) | -6.81 | 7.69 | 5.88 | 213.06 | 8.04 |
| PM10 ($\mu g/m^3$) | -4.36 | 18.08 | 15.23 | 260.31 | 13.38 |
| NEPH | -0.06 | 0.28 | 0.19 | 10.74 | 0.43 |
| DEM (m) – 5m resolution | -60.10 | 114.5 | 86.35 | 766.29 | 109.85 |
| DEM (m) – 30m resolution | 0.00 | 187.29 | 101.60 | 2,224.32 | 266.09 |
| Population density (persons/km$^2$) | 0.00 | 4,924.55 | 3,039.08 | 516,666.67 | 9,576.72 |
| Weekly personal median income | 0.00 | 845.99 | 805.00 | 3,500.00 | 320.29 |
| Land use mix entropy | 0.00 | 0.15 | 0.00 | 0.97 | 0.18 |

**Table 5.** Descriptive statistics of the input data for the cycling model.

| Variables | Min | Mean | Median | Max | Std. deviation |
| --- | --- | --- | --- | --- | --- |
| Observed cycling counts | 0.00 | 188.38 | 120.00 | 1,358 | 184.95 |
| Strava cycling counts | 0.00 | 12.33 | 5.00 | 6,535 | 50.04 |
| Precipitation (mm) | 0.00 | 3.16 | 0.00 | 206.00 | 10.67 |
| Temperature (C) | 8.80 | 23.20 | 22.70 | 47.10 | 5.20 |
| PM2.5 ($\mu g/m^3$) | -6.80 | 8.59 | 6.60 | 250.20 | 9.51 |
| PM10 ($\mu g/m^3$) | -5.70 | 19.21 | 16.00 | 268.6 | 14.98 |
| NEPH | -0.06 | 0.31 | 0.20 | 14.66 | 0.53 |
| DEM (m) – 5m resolution | -19.54 | 28.23 | 23.25 | 132.58 | 24.43 |
| Population Density (persons/km$^2$) | 0 | 4,959 | 3,367 | 318,333 | 7,533 |
| Weekly Personal Median Income | 0.00 | 744.10 | 718.00 | 3,000.00 | 293.34 |
| Land Use Mix Entropy | 0.00 | 0.14 | 0.00 | 0.80 | 0.19 |

## 3. MODELING APPROACH

### 3.1. Initial assessment of the original mobile phone and crowdsourced data

To better understand the error in the original mobile phone and crowdsourced data, we first assess both the mobile phone-based and Strava data's ability to directly explain observed walking and cycling volumes. Figure 3 shows the relationship between link-level mobile phone-based walking counts vs. observed walking counts across all count sites in the study area, as well as the relationship between link-level Strava cycling counts vs. observed cycling counts. Significant underestimation is observed across both data sources compared to the observed counts. Table 6 provides a summary of the performance of the two data sources against official observed walking and cycling counts. The analysis results confirm the presence of significant bias and under-representation issues in crowdsourced and mobile phone data.

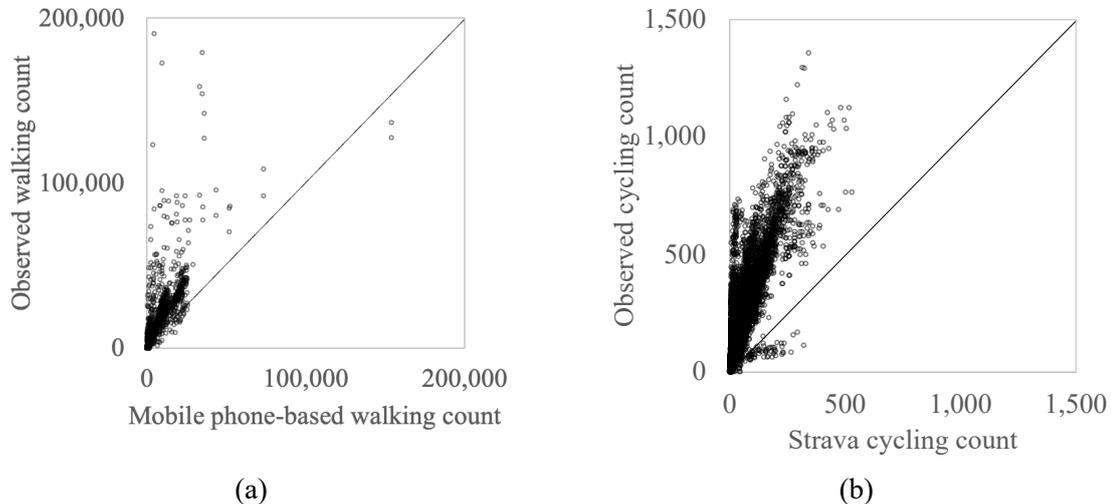

(a) (b)

**Figure 3.** (a) Relationship between the mobile phone-based walking counts and observed walking counts, and (b) Relationship between Strava cycling counts and observed cycling counts across all sites in the study area.

**Table 6.** Performance of the original mobile phone-based walking count and Strava cycling count data against observed walking and cycling counts across all sites in the study area.

| Data Source | $R^2$ | MAE | RMSE |
|---|---|---|---|
| Mobile phone-based (walking) | 0.38 | 5,483 | 12,906 |
| Strava (bicycling) | -0.17* | 147 | 201 |

* The best possible value of $R^2$ (coefficient of determination) is 1.0. A constant model that always predicts the average response variable regarding the input features gets a $R^2$ of 0.0. Negative $R^2$ suggests that the model is worse than the constant model.

### 3.2. Development of base (naïve) models

Next, we build the naïve or base models. Using the mobile phone-based walking count data as the sole predictor, an ordinary least squares (OLS) regression is estimated. We also estimate a similar OLS regression model using Strava cycling count data as the sole predictor. See Table 7 for the estimated coefficients and the models' performance metrics. Figure 4 also illustrates the estimated walking and cycling counts against the observed counts when the base models are used. Results suggest that on average, every single Strava bicycle trip in the study area represents 3.14 total cycling trips while for the mobile phone-based walking data every single counted pedestrian trip represents 1.78 total pedestrian trips. While the naïve models' performance exhibit significant improvement compared with the original crowdsourced and mobile phone data, the following sections of the manuscript demonstrate the extent that the models' performance can further improve with extensive machine learning training and testing, and when integrated with other data sources.

**Table 7.** Performance metrics for the base (naïve) models trained on the entire dataset without any held-out test.

| Model | β | $R^2$ | MAE | RMSE |
|---|---|---|---|---|
| Bicycling | 3.16 | 0.65 | 65 | 110 |
| Walking | 1.78 | 0.54 | 3,861 | 11,094 |

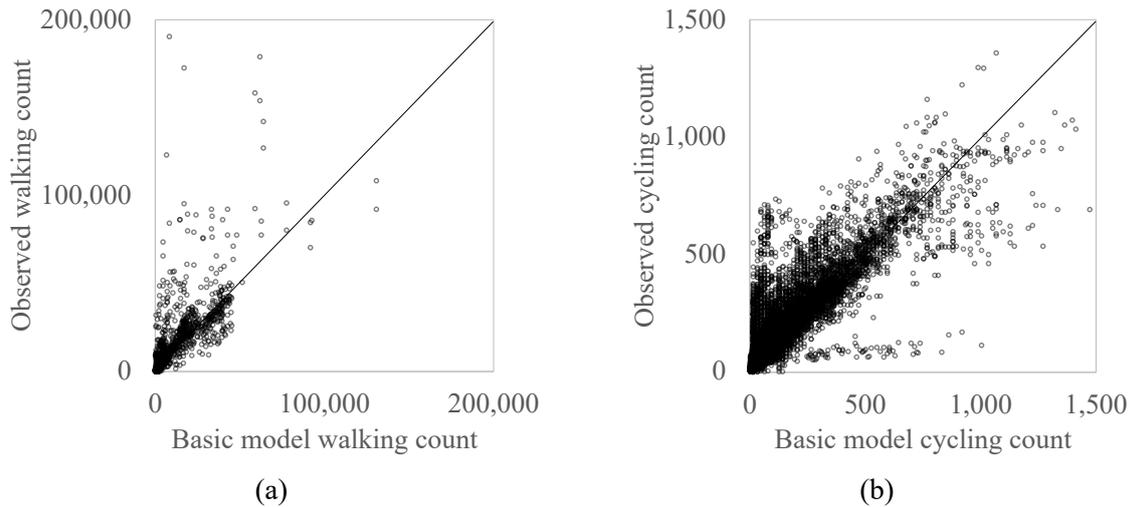

**Figure 4.** Base (naïve) (a) walking model and (b) cycling model count estimates against observed walking and cycling counts.

### 3.3. Feature selection and importance analysis

To identify which features should be further explored and included in the model training, we use domain-specific knowledge, insights offered from the published literature (Jestico et al., 2016; Chen et al., 2020; Nelson et al., 2021a) and a commonly used technique known as Least Absolute Shrinkage and Selection Operation (LASSO). Tables 8- 9 provide a summary of the LASSO analysis as well as estimated variance inflation factor (VIF) to assess variables multicollinearity across all the variables initially considered in the modeling. Generally, higher LASSO coefficient suggests a greater feature importance. VIF equal to one means variables are not correlated and multicollinearity does not exist in the model.

The LASSO analysis for the walking model reveals that the mobile phone-based walking count, population density, and Place of Interest (POI) density are the most influential features, while other variables such as the weather and climate indicators have lower importance. The VIF indicates low multicollinearity. For the cycling model, Strava cycling count, population density, and slope are crucial, with weekday/weekend classification and bicycle infrastructure also significant. COVID-related factors show moderate importance. VIF values suggest minimal multicollinearity. Overall, the findings guide feature selection, emphasizing key variables for model training and testing in the next section.

**Table 8.** Feature importance and variables multicollinearity analysis, ranked by highest to lowest LASSO coefficient, for the walking model.

| Input feature | LASSO | VIF |
|---|---|---|
| Mobile phone-based walking count | 11656.00 | 1.87 |
| Population density | 3742.60 | 1.17 |
| Place of Interest (POI) density | 3724.09 | 9.35 |
| Percentage of "walk linked" trips | 2426.60 | 65.84 |
| NO (nitric oxide) | 1791.71 | 2.59 |
| Minimum temperature | 1288.8 | 35.94 |
| Average slope | 1077.99 | 5.30 |
| $NO_2$ (nitrogen dioxide) | 607.45 | 8.33 |
| $O_3$ (ozone) | 604.51 | 10.90 |
| Parkland land use | 490.36 | 1.67 |
| Weekend/Weekday | 483.08 | 23.04 |
| PM10 | 481.84 | 1.58 |
| Median personal income | 479.18 | 22.79 |
| PM2.5 | 392.29 | 3.02 |
| Land use mix entropy | 387.19 | 19.52 |
| "Tertiary" links (OSM) | 359.06 | 1.19 |
| Precipitation | 346.97 | 1.57 |
| NEPH | 213.31 | 8.73 |
| "Residential" links (OSM) | 192.02 | 1.61 |
| Maximum temperature (C) | 32.19 | 81.2 |
| "Footway" links (OSM) | 25.33 | 3.70 |
| Maximum slope | 17.22 | 8.04 |

**Table 9.** Feature importance and variables multicollinearity analysis, ranked by highest to lowest LASSO coefficient, for the cycling model.

| Input feature | LASSO | VIF |
|---|---|---|
| Strava cycling count | 1358.41 | 1.82 |
| Population density | 236.53 | 1.93 |
| PM2.5 | 206.51 | 26.24 |
| NEPH | 115.49 | 14.93 |
| Average slope | 97.37 | 13.37 |
| PM10 | 95.08 | 18.34 |
| Land use mix entropy | 67.81 | 2.37 |
| COVID lockdown | 40.05 | 1.20 |
| Temperature | 21.74 | 14.56 |
| Precipitation | 21.27 | 1.26 |
| Weekday/Weekend | 16.84 | 2.98 |
| Maximum slope | 12.78 | 15.25 |
| Bicycle infrastructure | 10.68 | 1.78 |
| Pre-COVID time | 0.07 | 1.69 |
| Median personal income | 0.00 | 5.54 |

## 3.4. Final models' variable selection

Following an extensive model training, validation and testing the following variables are kept in the final walking and cycling models. See Table 10 and Table 11. The LASSO analysis confirms that the selected variables for the final models are of significant importance with high LASSO coefficient and no strong multicollinearity except a few variables in the walking model. As expected, the mobile phone-based walking count and Strava cycling count are the most important input feature in the walking and cycling models, respectively based on both LASSO and GINI coefficients. Generally, if VIF > 10, then multicollinearity is high. Note that multicollinearity does not reduce the predictive power or reliability of a model as a whole; it only affects estimations regarding individual predictors. In other words, a multivariable regression model featuring collinear predictors can provide insights into the overall predictive performance of the entire set of predictors concerning the outcome variable. However, it might not yield reliable findings regarding any specific predictor or identify which predictors overlap with others in terms of redundancy.

The only selected variables in the walking model that show signs of high multicollinearity include the percentage of "walk linked" trips and median personal income, as well as minimum temperature and maximum temperature. The percentage of "walk linked" trips and median personal income are likely to have correlation with population density and land use mix entropy in Sydney. Maximum and minimum temperatures are also highly correlated in a single day. Despite the existence of multicollinearity, we have kept the variables in the model as they do not affect the predictive power of the developed models.

**Table 10.** Final walking model variables: LASSO, GINI and VIF analysis (ranked by LASSO coefficient)

| Input feature | LASSO | VIF | GINI |
| --- | --- | --- | --- |
| Mobile phone-based walking count | 11874.54 | 1.80 | 0.74 |
| Place of Interest (POI) density | 4667.54 | 7.47 | 0.01 |
| Population density | 3853.82 | 1.160 | 0.12 |
| Percentage of "walk linked" trips | 2952.51 | 48.84* | 0.00 |
| Average slope (%) | 1219.64 | 3.68 | 0.04 |
| Weekend/Weekday | 806.44 | 1.44 | 0.00 |
| Median personal income | 791.38 | 22.61* | 0.01 |
| Precipitation | 641.50 | 1.45 | 0.01 |
| Parkland land use | 587.85 | 1.59 | 0.01 |
| Land use mix entropy | 528.41 | 2.79 | 0.03 |
| Minimum temperature (C) | 528.00 | 27.74* | 0.01 |
| "Tertiary" links (OSM) | 498.91 | 1.17 | 0.00 |
| Maximum temperature (C) | 370.81 | 67.94* | 0.00 |
| "Footway" links (OSM) | 299.13 | 3.27 | 0.00 |
| "Residential" links (OSM) | 201.79 | 1.50 | 0.00 |
| PM10 | 33.63 | 12.59* | 0.01 |

* Variables with high multicollinearity

**Table 11.** Final cycling model variables: LASSO, GINI and VIF analysis (ranked by LASSO coefficient)

| Input feature | LASSO | VIF | GINI |
|---|---|---|---|
| Strava count | 1626.82 | 1.62 | 0.71 |
| Population density | 198.50 | 1.62 | 0.08 |
| Maximum slope | 65.99 | 3.28 | 0.07 |
| Land use mix entropy | 50.01 | 1.65 | 0.06 |
| PM2.5 | 14.49 | 2.93 | 0.06 |
| Weekend/Weekday | 13.65 | 2.10 | 0.01 |
| Pre-COVID | 4.61 | 1.51 | 0.00 |

### 3.5. Spatial and temporal cross-validation

Cross-validation (CV) splits data into different sets to avoid model overfitting and bias toward known locations. For both the walking and cycling models, we perform a spatial split using count locations to separate the training set and the test set. We then perform a joint spatial and temporal split within the training set to perform a 10-fold CV and estimate 10 variations of the models. Benchmarks on all model variations are performed on the test set from the first split to evaluate the model performance.

The inclusion of spatial and temporal CV in model development, particularly in the presence of substantial spatial and temporal heterogeneity within observed data, holds significant importance for fortifying the model's robustness and applicability especially in large-scale networks. This methodological approach enables a more robust exploration of the inherent variations in diverse walking and cycling count data sampling locations and times. This rigorous evaluation safeguards the developed models against overfitting, enhances pattern discernment across locales and time frames, and serves as a vital quality control measure.

### 3.6. Model training and testing

Here, we comparatively examine, train, and test several statistical, machine learning and ensemble learning models including:

- Linear regression
- Support vector regression (SVR)
- Gaussian Naïve Bayes
- Tree regression
- Multi-layer perceptron (MLP)
- Adaptive Boosting (AdaBoost)
- Gradient Boosting regression
- Random Forest
- Voting regressor (incl. linear regression, MLP and Random Forest)
- Stacking regressor (incl. linear regression, Random Forest, Gradient Boosting, AdaBoost)

SVR is a machine learning algorithm that uses support vector machines to perform regression tasks. It aims to find a hyperplane that best fits the data while minimizing the error. Gaussian Naïve Bayes assumes that features are conditionally independent given the class label and follows a Gaussian (normal) distribution. Tree regression involves constructing a decision tree to model the relationship between features and the target variable in a regression task. MLP is a type of artificial neural network with multiple layers of nodes (neurons) organized into input, hidden, and output layers. It is widely used for classification and regression. Ensemble learning involves combining machine learning algorithms to address regression challenges. Ensemble learning strategically blends various ML models to enhance performance compared to a single model. AdaBoost is an ensemble learning method that combines the predictions of weak learners (often simple decision trees) to create a strong learner. It assigns weights to data points and focuses on misclassified samples to improve performance iteratively. Gradient Boosting Regression is also an ensemble learning technique that builds a series of weak learners (usually decision trees) sequentially. It aims to correct errors made by previous models by assigning more weight to misclassified observations. Finally, voting and stacking regressors are also types of ensemble learning models. Stacking aims to learn the optimal way to combine the predictions of the base models while Voting focuses on aggregating the predictions to make a final prediction.

To assess both the walking and cycling models' goodness of fit, we use three main measures of $R^2$, Mean Absolute Error (MAE) and Root Mean Square Error (RMSE) computed on both the training and testing data set as shown in Table 12 and 13. Figure 5 also provides a comparative illustration of the predictive power of a set of the developed walking and cycling models.

**Table 12.** Summary of the walking models' goodness of fit measures.

| Models | $R^2$ | MAE | RMSE | $R^2$ | MAE | RMSE |
|---|---|---|---|---|---|---|
| | Training | | | Testing | | |
| Linear Regression | 0.68 | 3,809 | 9,107 | 0.69 | 4,993 | 9,609 |
| SVR | 0.07 | 6,877 | 15,756 | 0.19 | 5,629 | 15,175 |
| Gaussian Naïve Bayes | 0.93 | 708 | 4,420 | 0.16 | 5,511 | 15,498 |
| Tree Regression | 0.97 | 579 | 3,060 | 0.77 | 3,254 | 6,907 |
| MLP | 0.71 | 3,315 | 8,627 | 0.70 | 4,530 | 9,418 |
| AdaBoost | 0.79 | 4,928 | 7,487 | 0.81 | 5,223 | 7,325 |
| Random Forest | 0.94 | 1,031 | 3,940 | 0.75 | 4,434 | 8,482 |
| Gradient Boosting Regression | 0.92 | 1,981 | 4,717 | 0.83 | 2,847 | 6,275 |
| Voting Regression | 0.80 | 2,881 | 7,310 | 0.82 | 4,094 | 7,161 |
| Stacking Regression | 0.81 | 2,870 | 6,910 | 0.79 | 3,398 | 8,991 |

**Walking model.** Linear Regression displays $R^2$ values of 0.68 and 0.69 for training and testing, respectively, with higher testing MAE and RMSE, suggesting potential overfitting. SVR exhibits a low $R^2$, indicating a weak fit, and potential overfitting is implied by considerably lower testing values. Gaussian Naïve Bayes shows a high $R^2$ for training but a lower $R^2$ for testing, signifying potential overfitting. In contrast, Tree Regression demonstrates high $R^2$ values for both training and testing, indicative of a good fit with low MAE and RMSE. The MLP model presents moderate $R^2$ values for both training and testing, with lower testing MAE and RMSE, suggesting improved generalization. AdaBoost yields relatively high $R^2$ values for both training and testing, with reasonable testing MAE and RMSE. Voting Regression and Stacking Regression exhibit reasonable $R^2$ values for both training and testing, with slightly higher testing MAE and RMSE. While Voting Regression performs well, Stacking Regression shows a slightly lower $R^2$ for testing, indicating potential overfitting. Overall, Random Forest, Gradient Boosting Regression, and Tree Regression emerge as top performers, but Voting and Stacking Regression offer more consistent performance across both training and testing dataset.

**Table 13.** Summary of the cycling models' goodness of fit measures.

| Models | $R^2$ | MAE | RMSE | $R^2$ | MAE | RMSE |
|---|---|---|---|---|---|---|
| | Training | | | Testing | | |
| Linear Regression | 0.71 | 67 | 100 | 0.83 | 52 | 65 |
| SVR | 0.36 | 96 | 137 | 0.56 | 120 | 174 |
| Gaussian Naïve Bayes | -2.26 | 259 | 309 | -0.41 | 279 | 312 |
| Tree Regression | 0.94 | 22 | 45 | 0.86 | 43 | 71 |
| MLP | 0.83 | 51 | 78 | 0.78 | 28 | 37 |
| AdaBoost | 0.75 | 67 | 85 | 0.85 | 85 | 103 |
| Random Forest | 0.94 | 25 | 45 | 0.87 | 45 | 69 |
| Gradient Boosting Regression | 0.88 | 37 | 60 | 0.86 | 63 | 98 |
| Voting Regression | 0.88 | 40 | 63 | 0.81 | 60 | 81 |
| Stacking Regression | 0.69 | 58 | 95 | 0.84 | 73 | 106 |

**Cycling model.** With the Linear Regression, the $R^2$ values for training and testing are 0.71 and 0.83, respectively. However, the MAE and RMSE values for testing are lower than for training, suggesting potential overfitting. The SVR model shows a moderate $R^2$ for both training and testing, with testing MAE and RMSE slightly higher than the training values. Gaussian Naïve Bayes exhibits negative $R^2$, indicating a poor fit. The Tree Regression model stands out with high $R^2$ values for both training and testing, suggesting an excellent fit, and low MAE and RMSE values. MLP presents high $R^2$ values for both training and testing, indicating effective generalization. AdaBoost and Random Forest show strong performance, with high $R^2$ values and reasonable testing MAE and RMSE. Gradient Boosting Regression, Voting Regression, and Stacking Regression demonstrate consistently good performance, maintaining high $R^2$ values for both training and testing, with relatively low MAE and RMSE values. In summary, for the cycling model, Tree Regression, MLP, AdaBoost, Random Forest, Voting Regression, and Gradient Boosting Regression stand out as top performers, with Voting Regression and Gradient Boosting Regression demonstrating the most consistent performance across both training and testing datasets. These models showcase a reasonable balance between a good fit and effective generalization.

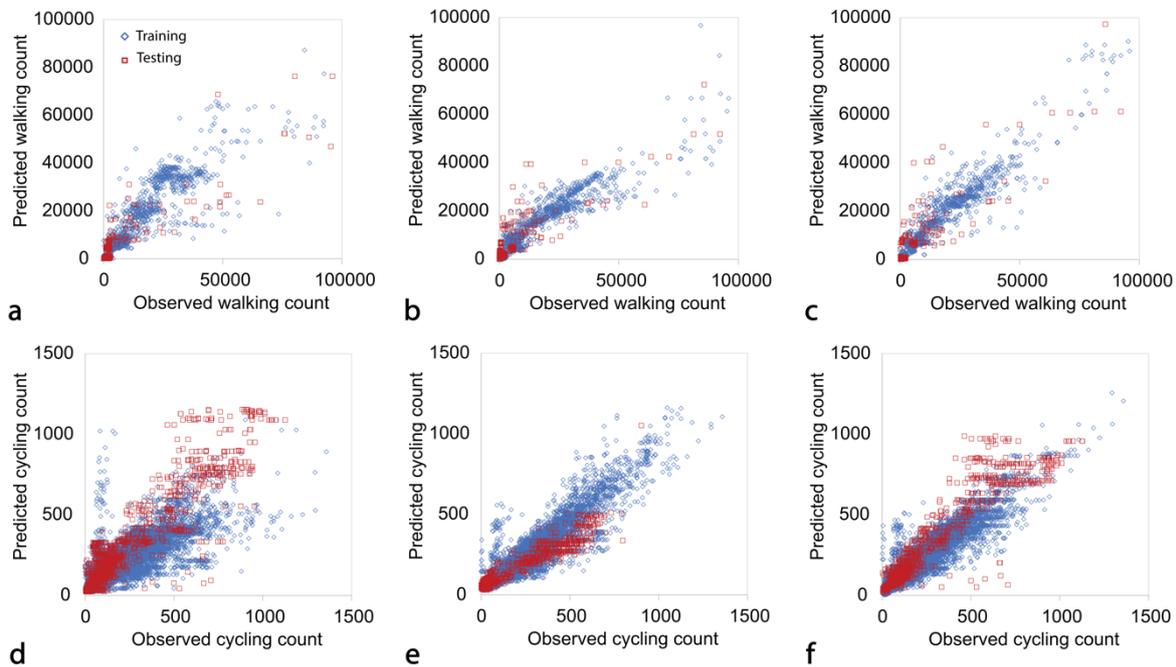

**Figure 5.** Illustration of the predictive power of selected developed walking models: (a) Stacking Regression, (b) Voting Regression, and (c) Gradient Boosting Regression; and cycling models: (d) Stacking Regression, (e) Voting Regression, and (f) Gradient Boosting Regression. Y axis represents the model predictions and X axis represents the observed walking or cycling counts across the study area. Blue dots represent the training test and red dots represent the test set.

## 4. MODEL INFERENCE AND APPLICATION

In this section, we apply the developed the stacking regression walking and cycling models, only as an example, to the entire NSW Six Cities Region to estimate the following outputs across the study area:

1. Average daily walking and cycling volumes at the link-level. See Figure 6.
2. Average daily walking and cycling kilometer travelled and per local government area (LGA).
3. Average number of walking and cycling trips per day and per LGA. See Figure 7.

We use link-level mobile phone-based walking count and Strava cycling count data from January 2023 to estimate and explore the spatial distribution of the walking and cycling trips across the study area.

To estimate the total number of walking and cycling trips in the study area, we aggregate and convert the link-level count estimates across all links as follows.

$$\text{Total number of walking trips} = \frac{\sum_i ADPV_i \cdot l_i}{d_w/\gamma_w} \quad (2)$$

$$\text{Total number of bicycling trips} = \frac{\sum_i ADBV_i \cdot l_i}{d_b/\gamma_b} \quad (3)$$

where $d_w$ and $d_b$ denote average walking and cycling trip distance, respectively. $\gamma_w$ and $\gamma_b$ denote effective walking and cycling distance factors, respectively. $ADPV_i$ represents the estimated Average Daily Pedestrian Volume for link $i$ while $ADBV_i$ represents the estimated Average Daily Bicycling Volume for link $i$. $l_i$ represents the length of link $i$. Note that $\sum_i ADPV_i \cdot l_i$ represents the total walking kilometers travelled and $\sum_i ADBV_i \cdot l_i$ represents the total bicycle kilometers travelled across the study area. Here, we assume that the average walking trip distance is 1 km based on the NSW HTS data and the average cycling trip distance is 4.7 km based on the Sydney Cycling Survey 2011 data. We also define the effective distance factor as the average percentage of a link length traversed by a pedestrian or cyclist. Given the short distance nature of most walking and cycling trips, the entire length of a link may not be entirely traversed by a pedestrian when walking from an origin or to a destination, especially when the walking and cycling trips are inferred through the motion of mobile phones. Here, we assume an effective distance factor of $\gamma_w = 0.5$ for the walking model and an effective distance factor of $\gamma_b = 1$ for the cycling model. Assuming a larger effective distance factor will increase the estimated total aggregated number of trips.

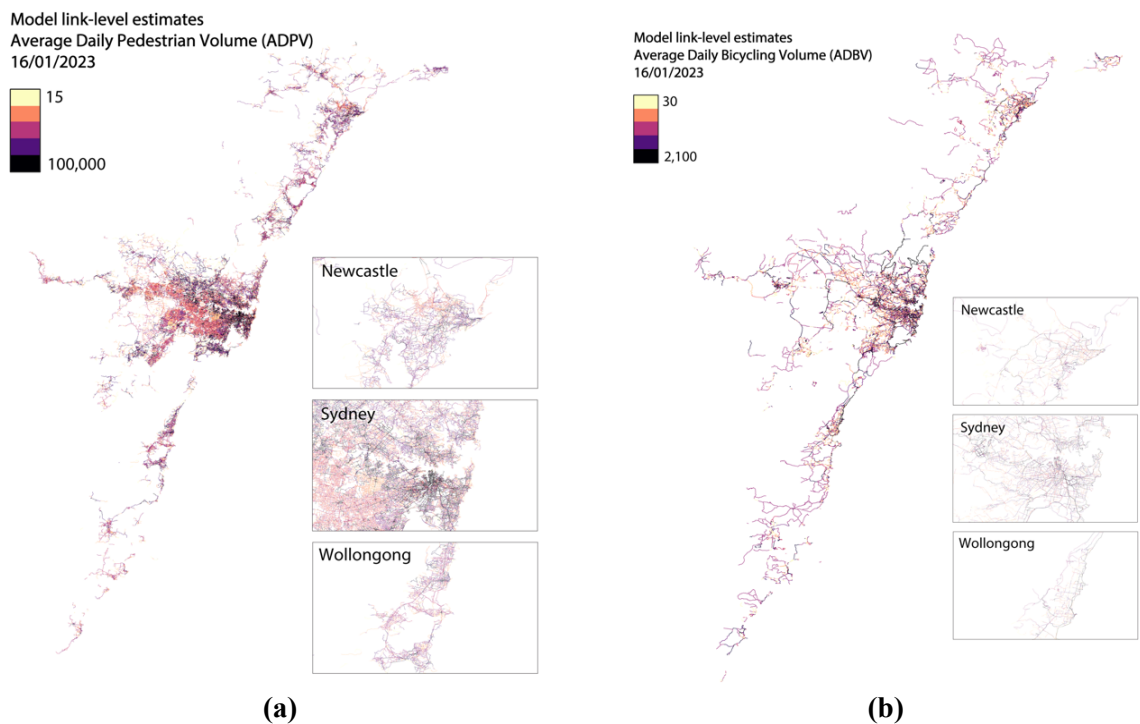

**Figure 6.** Estimated link-level (a) walking and (b) cycling trips across the NSW Six Cities Regional Network on 16/01/2023.

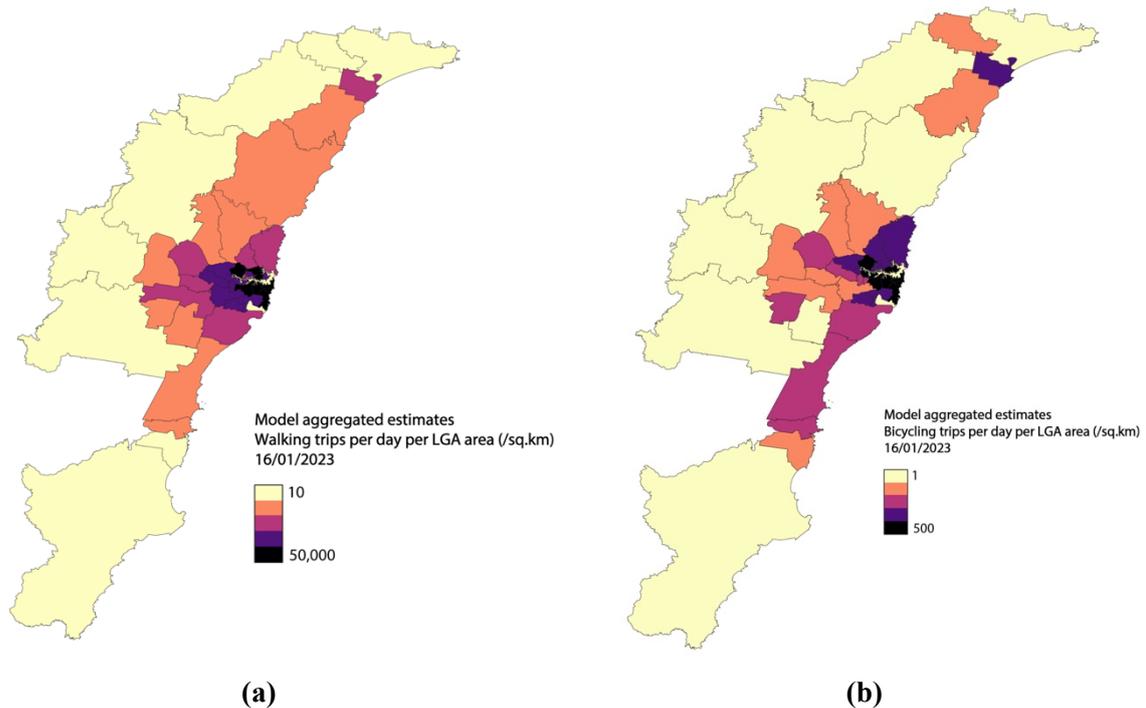

**(a)** **(b)**
**Figure 7.** Illustration of the estimated aggregated number of (a) walking trips and (b) cycling trips per LGA across the study area on 16/01/2023.

## 4.1. Identifying model estimation outliers

Given both walking and cycling models are trained mostly with data from relatively high pedestrian and cycling activity locations, the models tend to suffer from over-estimation in low pedestrian and cycling activity areas. To identify links with over-estimated walking and cycling counts, we adopt the "Kneedle" algorithm (Satoppa et al., 2011) to identify the outliers. The Kneedle algorithm is applied to the distribution of the ratio of the model estimates over the mobile phone-based walking counts or Strava cycling counts and systematically estimates the "elbow" point, estimated as 8 for the walking model and 16 for the cycling model. The estimated elbow point can be used as a cutoff line for outlier identification represented by the heavy tail of the distributions show in Figure 8.

The estimated elbow points are used to identify links in the network with excessive over-estimation. With the estimated elbow point of 8 for the walking model, about 10% of the link estimates across the entire study area are identified as outliers. While with the estimated elbow point of 16 for the cycling model, only less than 6% of the link estimates across the network are identified as outliers. The analysis suggests that while the developed walking model suffers from over-estimation, the cycling model estimates are more reasonable. The overestimation evident in the walking model signifies a tendency towards amplifying pedestrian counts, particularly in locales characterized by minimal pedestrian activity. This tendency introduces a susceptibility to inaccuracies in predictive outcomes. The cycling model, however, exhibits more robust estimations, indicative of enhanced reliability in forecasting cycling counts across a spectrum of locations. This difference emphasizes the need to be cautious when using the walking model in areas with low pedestrian activity, while the cycling model proves to be a more reliable option for various situations.

A few practical solutions to tackle the issue of the over-estimation in the walking model include applying a hard upper bound constraint on the ratio of the model walking count estimates over the mobile phone-based walking counts. An alternative approach is to replace the estimation of the walking counts from the proposed stacking regression model with the base model (presented earlier in Table 7). This would mitigate the impact of the over-estimation issue for the identified subset of links across the network without affecting the model estimations in areas where over-estimation is not present or not as significant that is about 82% of the links in the network.

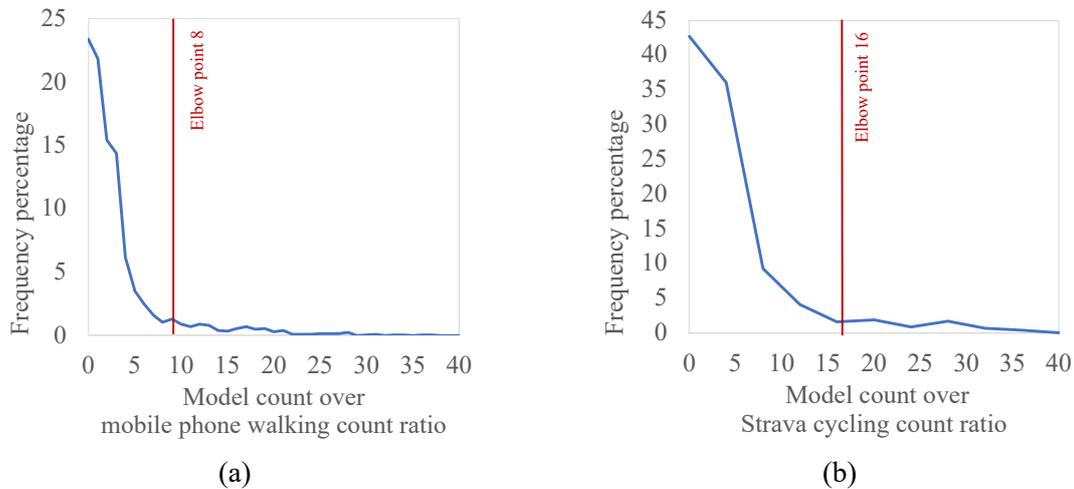

**Figure 8.** Distribution of the (a) ratio of the estimated link-level walking counts over the mobile phone-based walking counts and (b) ratio of the estimated link-level cycling counts over Strava cycling counts.

### 4.2. Comparison against the NSW HTS data

The NSW government's latest estimates of the total number of active transport trips across the NSW state is 2.1 billion trips per year including 1.5 billion walking and bicycling trips with 600 million additional active transport trips connected to public transport. Our modelling estimates suggest a total of 2.6 billion active transport trips (including both walking and bicycling) per year occur across the NSW Six Cities Region.

The NSW HTS data suggests that across the Sydney Greater Metropolitan Area (GMA) which includes Sydney Greater Capital City Statistical Area (GCCSA), parts of Illawarra and Hunter regions (including a total of 44 LGAs), 7.74 million walking trips occur per day without distinguishing between weekdays and weekends. Note that the NSW HTS data does not include tourists or visitor trips and is only limited to NSW residents. The developed walking model estimates that a total of 7.09 million walking trips occur per weekday and 6.67 million walking trips occur per weekend in the NSW Six Cities region.

The NSW HTS data does not provide the number of cycling trips as a separate individual mode. Instead, it suggests 440,000 "other" trips per day that includes cycling, ridesharing, carsharing, etc. Optimistically assuming 50% of all the "other" trips in the HTS are associated with cycling, there would be 220,000 cycling trips per day occurring across the study area. This estimate is comparable with the number of cycling trips per day estimated by our cycling model. Our developed cycling model estimates that a total of 260,000 cycling trips occur per weekday and 406,000 cycling trips occur per weekend in the NSW Six Cities region.

### 5. DISCUSSION

Quality and quantity of input data are two critical elements in any statistical and machine learning modelling. Building upon the numerical results presented in the earlier section, here we provide a thorough discussion on some of the major challenges and limitations of the presented modeling framework and recommended future research directions.

**Links with no data.** Strava cycling data does not include links with number of cycling trips smaller than 5 per day. This potentially creates many links with no data in any single day that may have low cycling activity. A potential solution to mitigate the impact of links with no data is to look at a longer time period (for example one year's worth of Strava raw data) and take an average Strava cycling count for each link. Same limitation applies to the mobile phone-based walking count data in which links with very low walking volume are not included and reported because of larger uncertainty in the estimates. This uncertainty essentially carries over to any model estimation based on mobile phone and crowdsourced data.

**Observed count data.** The number of locations with observed walking and cycling count data were limited in our study. This affects the predictive power of the developed models and can introduce uncertainty in model estimations specially for areas with no observed count data. Also, the locations with observed walking and cycling count data were not geographically diverse. Most official count locations were from high pedestrian and cycling activity areas (e.g. City of Sydney and City of Parramatta) that naturally create a bias in the model outcomes. Therefore, future model development and improvement should try to maximize the geographical diversity and temporal coverage of the official count data locations as much as possible. Given both the walking and cycling models were trained and tested on limited observed count data from relatively high walking and cycling activity areas, both models tend to overestimate the number of walking and cycling trips in low activity areas.

**Link-level volume inconsistencies.** Link-level estimation of walking and cycling volumes could create potential inconsistencies in some streets or corridors that are made of multiple smaller links where the smaller links have different attributes such as slope or associated population or land use characteristics. To address this limitation, uncertainty bounds around the estimated mean volumes could be estimated.

While mobile phone and crowdsourced data (e.g. Strava and mobile phone-based) may offer insights on some major walking and cycling corridors, when the link-level model estimates are applied, a uniform normalization should not be expected across the space (e.g. route or corridor). For example, a corridor could be a major route for recreational cycling trips given Strava user types but when the developed model normalizes the Strava raw data to the whole population, some parts of the corridor depending on its land use and population characteristics may not remain a major route for the whole population. Thus, the model outcomes may no longer presents similar distinct corridors and routes as in the Strava or the mobile phone-based base data.

**Network size.** The impact of the network size on the model accuracy at micro scales could be significant. A model trained and validated for the entire NSW Six Cities Region performs differently compared to a much smaller model just trained and validated for a single LGA only. One approach to address this limitation is to develop different models for different regions (urban vs. regional) as opposed to a single model for the entire study area. However, each approach has its own pros and cons. In an extreme case, one could potentially build a model per LGA but this wouldn't be much practical from an operational and model maintenance perspective. Extension of the developed models to ever larger area such as the entire NSW state should not be applied without proper re-training and re-testing of the models to the regional areas in the state. Otherwise, the model outcomes will likely suffer from significant over-estimation in regional areas.

**Spatial aggregation.** The use of SA1 and LGA geographical boundaries for some of the input features may create spatial biases. To address this limitation, use of Modifiable Areal Unit Problem (MAUP) could be explored and tested. However, we highlight that the use of SA1 and LGA boundaries is a common practice in many transport planning and modelling applications mainly because of the consistency that it creates across different data sets and model inputs and outputs.

**Need for model re-training and re-testing.** The walking and cycling models should ideally be re-trained and re-tested periodically with new observed walking and cycling count data across as many locations as possible in the study area if intended to use in practice for active transportation planning and operations.

**Operational vs. strategic modeling.** The developed walking and cycling models are purely data-driven and for operations applications rather than long-term planning. The developed models do not include any behavioral theory capturing travelers' behavior, route and destination choices, and user preferences.

An interesting direction for future research is development of a strategic active transportation network planning model empowered by a range of behavioral demand and supply models.

**ACKNOWLEDGEMENT**

Authors acknowledge Transport for NSW (Cities and Active Transport) for providing access to data, information, and insights. We are also grateful for the support and contributions from Sandeep Mathur, Tony Arnold, and Mohammed Rashid. Additionally, the authors thank Ben Beck, Chris Pettit, Parisa Zare, and Trisalyn Nelson for their contributions during the early stages of the project.